\documentclass{article}

\usepackage{mwe}
\usepackage{xr}
\usepackage{microtype}
\usepackage{graphicx}
\usepackage{booktabs}
\usepackage{hyperref}
\gdef\isaccepted{1} 
\usepackage{icml2020}
\usepackage{comment}
\usepackage{hyperref}
\usepackage{natbib}
\usepackage{amsmath}
\usepackage{amsfonts}
\usepackage{bbm}
\usepackage{mathtools}
\usepackage{algorithm}
\usepackage[noend]{algpseudocode}
\usepackage{url}
\usepackage{graphicx}
\usepackage{subcaption}
\icmltitlerunning{Universal Value Density Estimation for Imitation Learning and Goal-Conditioned Reinforcement Learning}

\makeatletter
\newcommand{\removelatexerror}{\let\@latex@error\@gobble}
\makeatother

\begin{document}

\twocolumn[
\icmltitle{Universal Value Density Estimation for Imitation Learning and Goal-Conditioned Reinforcement Learning}



\icmlsetsymbol{equal}{*}

\begin{icmlauthorlist}
\icmlauthor{Yannick Schroecker}{gt}
\icmlauthor{Charles Isbell}{gt}
\end{icmlauthorlist}

\icmlaffiliation{gt}{College of Computing, Georgia Institute of Technology}
\icmlcorrespondingauthor{Yannick Schroecker}{yannickschroecker@gatech.edu}

\icmlkeywords{Imitation Learning, Reinforcement Learning}

\vskip 0.3in
]

\printAffiliationsAndNotice{}  

\begin{abstract}
    This work considers two distinct settings: imitation learning and goal-conditioned reinforcement learning. In either case, effective solutions require the agent to reliably reach a specified state (a goal), or set of states (a demonstration). Drawing a connection between probabilistic long-term dynamics and the desired value function, this work introduces an approach which utilizes recent advances in density estimation to effectively learn to reach a given state. As our first contribution, we use this approach for goal-conditioned reinforcement learning and show that it is both efficient and does not suffer from hindsight bias in stochastic domains. 
    As our second contribution, we extend the approach to imitation learning and show that it achieves state-of-the art
    demonstration sample-efficiency on standard benchmark tasks.
\end{abstract}
\section{Introduction}\label{sec:introduction}
Effective imitation learning relies on information encoded in the demonstration states. In the past, successful and
sample-efficient approaches have attempted to match the distribution of the demonstrated
states~\citep{ziebart2007maximum,ho2016generative,schroecker2019generative}, reach any state that is part of the
demonstrations~\citep{wang2019random,reddy2019sqil}, or track a reference
trajectory to reproduce a specific sequence of states~\citep{peng2018deepmimic,aytar2018playing,pathak2018zeroshot}.
Without explicitly trying to reproduce demonstrated states, imitation learning suffers from the problem of accumulating
errors~\citep{ross2011reduction} and requires a larger amount of demonstration data to accurately reproduce the expert's
behavior. Furthermore, without reasoning about how to reach desired states, the agent will be unable to learn from observation alone.
The question of how to reach desired target states has also been separately considered as part of another field of study:
goal-conditioned reinforcement learning. Goal-conditioned reinforcement learning aims to train flexible  
agents that can solve multiple variations of a
given task by parameterizing it with a goal. Often, this goal takes the form of a state or desired observation that the
agent has to visit or achieve in the most optimal way. In this case, the task becomes highly similar to the setting of
imitation learning: where goal-conditioned reinforcement learning trains the agent to reach a single given goal state,
common approaches to efficient imitation learning teach the agent to reach multiple given goal states, i.e. to match
a distribution or sequence of demonstrated states. 

Despite significant achievements in the field~\citep{schaul2015universal,andrychowicz2017hindsight,nair2018visual,sahni2019visual}, learning to achieve arbitrary goals remains an extremely difficult
challenge.  In the absence of a suitably shaped reward function, the signal given to the agent can be as little as a
constant reward if the goal is achieved and 0 otherwise. Such a reward function is sparse and difficult to learn from,
a problem that is only exacerbated when the task is to achieve any arbitrary goal. 
Hindsight Experience Replay (HER)~\citep{andrychowicz2017hindsight} introduces the concept of hindsight samples as an
effective heuristic to tackle this problem. Hindsight sampling selects transitions from past experience but artificially
changes the goal during the learning process to pretend that the agent intended to reach the state that it actually observed later in the
roll-out. This way,
the agent will frequently observe a reward and receive a comparatively dense learning signal. HER provides remarkable speed-ups and is
capable of solving a variety of sparse, goal-conditioned RL problems; however, the approach cannot be applied to all
domains as it suffers from hindsight bias. 
By changing the goal in hindsight, the agent implicitly discards unsuccessful attempts.
As a result, if an action has a large failure rate and leads the agent to succeed
in only a small fraction of
all attempts, the value of the action will be
dramatically overestimated. 

Addressing problems in goal-conditioned reinforcement learning and in imitation learning, our work makes two central contributions: \begin{enumerate}
    \item We introduce an alternative and unbiased approach to utilize hindsight samples, enabling
        sample-efficient goal-conditioned reinforcement learning in domains that prior methods cannot solve.
    \item We introduce an imitation learning method based on this approach and show that it 
         outperforms the current state-of-the-art in demonstration sample-efficiency on common benchmark tasks.
\end{enumerate}

Utilizing hindsight samples alone to achieve unbiased goal-conditioned reinforcement learning in the general case is impossible: if we never sample negative transitions, we cannot train the agent to accurately assess the
risks of such transitions; however, the most common formulation and a useful special case assigns a positive reward to states in which the
goal has been achieved and provides a reward of 0 otherwise\footnote{This scenario is also the primary focus of the original Hindsight Experience
Replay experiments} (with an additional, optional term handling action cost). 
We observe that the long-term expected
reward in such a scenario is directly proportional to the discounted likelihood of achieving this goal. We propose the term ``Value Density'' to refer to a special case of a value function that is also a valid
representation of this likelihood. 
We furthermore observe that hindsight samples provide us with exactly the state-action and achieved goal triplets that 
are required for estimating the density of achieved goals. This gives rise to value density estimation: using a modern
density estimator~\citep{dinh2016density}, we utilize hindsight samples to directly estimate the value function.
Combined with regular temporal difference learning updates, this approach allows us to estimate the value function for
each goal in a way that is sample-efficient, unbiased and low-variance. We will explore this approach in detail in
Section \ref{sec:uvd_gcrl}.

To extend this approach to imitation learning, we teach the agent to reproduce the distribution of states that the expert teacher
has demonstrated to the agent. This encourages the agent to stay close to demonstration states and prevents
errors from accumulating, a problem which is largely responsible for inefficiencies in imitation learning approaches.
Errors accumulate when the agent deviates from demonstrations and has to make decisions in states that are unlike any it has seen
as part of the demonstration set. Sampling demonstration-states as goals, we can utilize the same approach to teach the
agent to stay near
the demonstrated states. To ensure that the agent attempts to visit all states equally often,
i.e. to ensure that the agent matches the expert's state-distribution rather than sticking to a subset of demonstrated
states, we have to pick demonstration states as goals with higher probability if the state is unlikely to be visited by
the agent and vice versa. To this end, we maintain a model of the agent's state-distribution as well. We describe this
approach in detail
in Section \ref{sec:vdi}.

\section{Background}
\subsection{Markov Decision Processes}\label{sec:mdp}
Markov Decision Processes (MDPs) are an essential formalism to describe sequential decision making problems such as
    reinforcement learning and imitation learning. Here, we briefly lay out notation while referring the reader to
    \citet{puterman2014markov} for a detailed review. MDPs define a set of states $\mathbb{S}$, a set of actions $\mathbb{A}$, a
    distribution of initial states $d_0(s)$, Markovian transition dynamics defining the probability
    (density) of transitioning from state $s$ to $s'$ when taking action $a$ as $p(s'|s, a)$, and a
    reward function $r(s,a)$. In reinforcement learning, we commonly wish to find a parametric stationary policy
    $\mu_\theta: \mathbb{S} \rightarrow \mathbb{A}$. Here, we write the policy as a deterministic function as we will utilize deterministic policy
    gradients, but all findings hold for stochastic policies as well. An optimal policy is one which maximizes the
    long-term discounted reward $J^r_\gamma(\theta)=E\left[\sum_{t=0}^\infty \gamma^t
    r(s_t, a_t)|s_0 \sim d_0, \mu_\theta \right]$ given a discount factor $\gamma$ or, sometimes, the average reward
    ${J^r(\theta)=\int
    d^{\pi_\theta}(s)r(s,\mu_\theta(s))ds,a}$, where the stationary state distribution $d^\mu(s)$ and the stationary
state-action distribution $\rho^\mu(s, a)$ are uniquely induced by
$\mu$ under mild ergodicity assumptions. A useful concept to this end is the value function
$V^\mu(s)=E\left[\sum_{t=0}^\infty \gamma^t r(s_t, a_t) |s_0=s, \mu\right]$ or Q function $Q^\mu(s, a)=E\left[\sum_{t=0}^\infty \gamma^t
    r(s_t, a_t) |s_0=s, a_0=a, \mu\right]$ which can be used to estimate the policy gradient $\nabla_\theta
    J_\gamma(\theta)$ \citep[e.g.][]{sutton1999policy,silver2014deterministic}.
    Finally, we define as $p_\mu(s,a \xrightarrow{t} s')$ the probability of transitioning from state $s$ to $s'$ after
    $t$ steps when taking action $a$ in state $s$ and following policy $\mu$.

    \subsection{Goal-conditioned Reinforcement Learning}\label{sec:gcrl}
    Goal-conditioned Reinforcement Learning aims to teach an agent to solve multiple variations of a task, identified by a goal vector $g$. Conditioned on the goal, the reward function $r^g(s,
    a)$ describes all possible instantiations of the task. To solve each possible variation, the agent learns a representation of a goal-conditioned policy, which we write as $\mu_\theta^g(s)$.
    Based on the goal-conditioned reward and policy, we can write down a generalized definition of
    the value function. Note that the policy and the reward can be
    conditioned on different goals:
    \begin{equation}
        V_{r^g}^{\mu^{\overline{g}}}(s)=E\left[\sum_{t=0}^\infty \gamma^t r^g(s_t, a_t) |s_0=s, \mu^{\overline{g}}\right]
    \end{equation}

    To solve such tasks, \citet{schaul2015universal} introduce the concept of a Universal Value Function Approximator (UVFA), a learned model
    $V_\omega(s; g)$ approximating $V_{r^g}^{\mu^{g}}(s)$, i.e. all value functions where the policy and reward are
    conditioned on the same goal. For the purposes of this work, we also consider models which generalize over different
    value-functions, but consider a fixed, specific policy. To distinguish such models visually, we write
    $\tilde{V}_\omega(s; g)$ to refer to models which approximate $V_{r^g}^{\mu}(s)$ for a given $\mu$. Where $V_\omega$
    represents how good the agent is at achieving any goal if it tries to achieve it, $\tilde{V}_\omega$ models how good a
    specific policy is, if the task were to achieve the given goal.

    \citet{schaul2015universal} show that UVFAs can be trained via regular temporal difference learning with randomly sampled
    goals. This allows the agent to learn a goal-conditioned policy using regular policy gradient updates; however, sampling goals at random requires the reward signal to be sufficiently dense. 
    A common use-case for goal-conditioned reinforcement learning involves solving problems with a sparse reward, for example an
    indicator function that tells the agent whether a goal has been achieved. In this case, the agent rarely
    observes a non-zero reward while training the UVFA.
        Hindsight updates (HER) \cite{andrychowicz2017hindsight} are a
        straight-forward solution, changing goals recorded in a replay memory based on what the agent has actually
        achieved in hindsight. While the approach is intuitive, it is also biased. We will explore the significance of this further in
        Section~\ref{sec:not_scaffolding} where we also introduce an alternative solution that is both efficient and
        unbiased. 
\subsection{Imitation Learning}\label{sec:imitation}
    Imitation learning (IL) teaches agents to act given demonstrated example behavior. While the MDP formalism can still be used in this
    scenario, we no longer have a pre-defined reward function specifying the objective. Instead, we are given a sequence of
    expert state-action pairs. The goal of imitation learning is to learn a policy
    $\mu_\theta$ that is equivalent to the expert's policy $\mu^*$ which generated the demonstrated states and actions.
    This problem is underspecified and different formalizations have been proposed to achieve this goal. 
    The simplest solution to IL is known as Behavioral Cloning (BC)~\citep{pomerleau1989alvinn} and treats the problem as a supervised learning problem. Using demonstrated states as sample inputs and demonstrated actions as target outputs, a policy can be
    trained easily without requiring further knowledge of, or interaction with, the environment. While this approach can
    work remarkably well, it is known to be suboptimal as a standard assumption of supervised learning is violated:
    predictions made by the agents affect future inputs to the policy~\citep{ross2011reduction}. By learning about the environment dynamics and
    explicitly trying to reproduce future demonstrated states, the agent is able to learn more robust policies from
    small amounts of demonstration data. 

    A common and effective formalism is to reason explicitly about matching the distribution of state-action pairs that the agent will
    see to that of the expert. By learning from interaction with the environment and employing sequential reasoning,
    approaches such as maximum entropy IRL~\citep{ziebart2007maximum}, adversarial IRL~\citep{fu2018learning}, GPRIL~\citep{schroecker2019generative} or Generative Adversarial Imitation Learning (GAIL)~\citep{ho2016generative} lead the agent to reproduce the expert's
    observations as well as the expert's actions. By drawing a connection to goal-conditioned reinforcement learning, we
    introduce a novel approach to match the expert's state-action distribution and will show that it outperforms the
    current state-of-the-art on common benchmark task. 
    \subsection{Normalizing Flows}\label{sec:normalizing_flows}
    Recent years have seen rapid advances in the field of deep generative modeling. 
    While much of the field has focused on generating representative
    samples~\citep{kingma2013autoencoding,goodfellow2014generative}, methods such as autoregressive
    models~\citep{vandenoord2016pixel,vandenoord2016wavenet} and normalizing
    flows~\citep{van2017parallel,dinh2016density} have demonstrated the ability to learn an explicit representation of complex density functions. 
    In this work, we require the ability to estimate highly non-linear Value Density functions and utilize a simplified version of
    RealNVP~\citep{dinh2016density} for this purpose. As a Normalizing Flow, RealNVPs consist of a series of learned, invertible transformations
     which transform samples $z\in \mathbb{R^N}$ from one distribution $p_z$ to samples $x\in \mathbb{R^N}$ from another distribution $p_x$. Chaining
    multiple bijective transformations allows the model to transform simple density functions, such as a unit Gaussian,
    to represent complex distributions such as images (in the original paper) or the agent's state (in this work).
    Normalizing flows represent the density function explicitly and in a differentiable way, such that the model can be
    trained via maximum likelihood optimization. To ensure that the density function and its gradient are tractable, special care has to
    be taken in picking the right bijective transformation. RealNVP utilizes affine transformations where
    half of the input features are scaled and shifted, with scale and shift parameters predicted based on the other half. While the original work defines a specific autoregressive order to model images effectively, we use a simplified version of RealNVP
            in this paper to model non-image data and pick the autoregressive order at random. 

\section{Universal Value Density Estimation}\label{sec:uvd}
\subsection{Addressing Hindsight Bias}\label{sec:not_scaffolding}
\begin{figure}
    \centering
    \centering
    \begin{subfigure}{0.2\textwidth}
        \includegraphics[width=\textwidth]{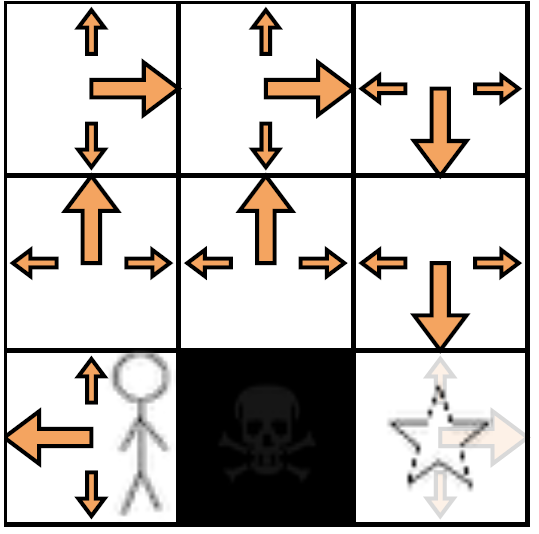}
        \caption{}
    \end{subfigure}
    \begin{subfigure}{0.2\textwidth}
        \includegraphics[width=\textwidth]{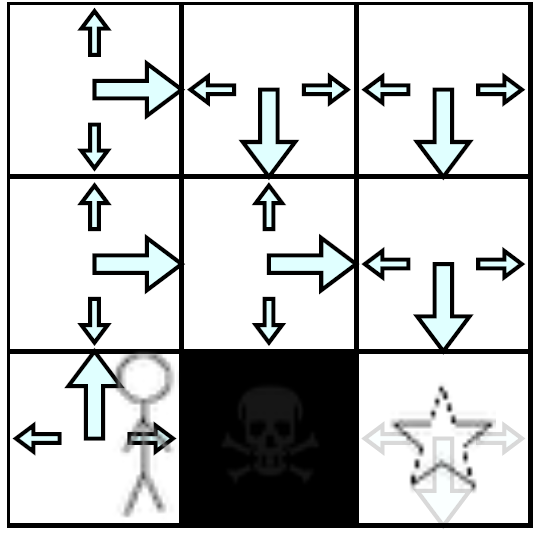}
        \caption{}
    \end{subfigure}
    \caption{Simple cliff-walk domain. The agent starts in the bottom left, has to avoid the cliff depicted in black and
reach the state in the bottom right. \textbf{a)} depicts the transition probabilities using an optimal policy while
\textbf{b)} depicts a sub-optimal policy learned using HER.}\label{fig:cliffwalk}
\end{figure}

To teach the agent to reach desired states, either for the purposes of goal-conditioned reinforcement learning or for
imitation learning, we first consider the question of learning approximations of the goal-conditioned value-functions
$\tilde{V}_\omega$ or $V_\omega$, using fixed or goal-conditioned policies respectively. Either requires us to address
the challenge of learning from sparse rewards.
\citet{andrychowicz2017hindsight} train UVFAs efficiently by
replacing the original goal with the achieved goal. Allowing the agent to learn from failures, HER provides an intuitive
way to speed up the training of a UVFA; however, as can be seen in Figure \ref{fig:cliffwalk}, the approach suffers from
hindsight bias. In this gridworld-example, the agent starts in the bottom left and
has to walk around a cliff to reach the goal in the bottom right. Environment noise can make the agent move
perpendicular to the desired direction and the optimal policy has to avoid transitions that might accidentally lead the
agent down the cliff. Using HER, we alter the goal on unsuccessful roll-outs and only learn about the true goal based on
successful ones. The agent never learns from transitions leading down the cliff, underestimates their probability and
chooses a shorter, but sub-optimal path.

To identify the source of hindsight bias, we can examine the effect of changing each sampling distribution in the
hindsight temporal difference update rule. Using $\overline{\omega}$ to refer to the parameters of the previous
iteration, the temporal difference update rule looks as follows: \begin{equation}\label{eq:UVFA_TD} 
    \begin{aligned}
    &\omega \leftarrow \overline{\omega} + \alpha \Delta \\ &\Delta  := \int \rho^{\mu_\theta^g}(s, a) p(s'|s, a) p(g) \nabla_\omega Q_\omega(s, a; g)\delta
    ds, a, s', g\\
    &\delta := r^g(s, a) + \gamma
    Q_{\overline{\omega}}(s', \mu^g_\theta(s')) - Q_{\overline{\omega}}(s, a).
    \end{aligned}
\end{equation}
The update rule performs a regression step, minimizing the distance between the Q-value of the state-action pair $s,a$
and the temporal difference target induced by the environment dynamics and the reward function. Altering
$\rho^{\mu_\theta^g}(s, a)$ is akin to using out-of-distribution samples in regression and is often done successfully in practice, for example by
using a replay buffer~\citep{munos2016safe, fujimoto2018addressing}. Altering the distribution of
goals $p(g)$ has a similar effect. These distributions, however, have to be independent from the calculation of the regression target.
Hindsight samples violate this assumption which in turn leads to hindsight bias. Sampling $s', g$ from $p(s'|s, a)p(g|s, a, s')$
is identical to sampling from $p(g| s, a)p(s'| s, a, g)$, i.e. the agent will underestimate the
probability of failure because the sampling procedure is equivalent to altering the transition dynamics such that they are more likely to lead to the goal.
\subsection{Value Density Estimation}
We consider the special case where the task is for the agent to reach a valid goal state. This scenario is
common in goal-conditioned RL and has also been considered by \citet{andrychowicz2017hindsight}. In
discrete environments, we can define such tasks by a reward that is positive if the goal is achieved and 0 otherwise,
we define: \[r^g(s,
a):=(1-\gamma)\mathbbm{1}(h(s, a)=g),\] where $h$ is a function that defines the achieved goal for any given state-action pair. In
slight abuse of notation\footnote{Formally, the reward would have to be defined to be non-zero only in an $\epsilon$-ball around $h(s,a)$ and inversely
proportional to the volume of this ball. All results hold in the limit $\epsilon\rightarrow 0$.}, we extend this
definition to continuous environments: \[r^g(s, a)=(1-\gamma)\delta_{h(s, a),
g}\]  

We can now show that the Q-function of such tasks forms a valid density function. Specifically, we notice  that the Q function
is equivalent to the discounted probability density over goals that the agent is likely to achieve when following its policy, starting from the given state-action pair:
\begin{equation}
\begin{aligned}\label{eq:density_def}
    Q^{\mu}_{r^g}(s, a) &= \mathbb{E}\left[\sum_{t=0}^\infty \gamma^t r(s_t, a_t)| s_0=s, a_0=a, \mu\right]\\
    &= (1-\gamma)\sum \gamma^t \int p^\mu(s, a \xrightarrow{t} s') \delta_{h(s', \mu(s')), g}
ds'\\
    &=: F^{\mu}_\gamma (g|s, a) 
\end{aligned}
\end{equation}

It follows that we can learn an approximation of the goal-conditioned Q-function $Q^\mu_{r^g}(s, a)$ by approximating
the value density $F_\gamma^\mu(g|s, a)$. This can be done using modern density estimators such as RealNVPs~(see Section
\ref{sec:normalizing_flows}). To train the model, we gather training samples from a roll-out $s_0, a_0, s_1, a_1,
\dots$, collecting state-action pairs $s=s_t, a=a_t$ at random time-steps $t$ as well as
future achieved goals $g=h(s_{t+j}, a_{t+j}); j\sim Geom(1-\gamma)$. Notice that the above derivation assumes a fixed,
i.e. not goal-conditioned policy. We relax this assumption in Section~\ref{sec:uvd_gcrl}.

\begin{algorithm}[t]
\begin{algorithmic}[1]
    \small
    \Function{UVD}{}
    \For{$i \gets 0..\text{\#Iterations}$}
        \State Fill replay buffer with experience
        \For{$s, a, \overline{g}$ sampled from short replay buffer}
            \State Sample time offsets $t\sim \text{Geom}\left(1 - \gamma\right)$
            \State Sample achieved goals $g$ $t$ steps ahead of $s$
            \State Update $F_\Phi$ with $-\nabla_\Phi \log F_\Phi\left(g|s, a, \overline{g}\right)$
        \EndFor

        \For{$s, a, s', \overline{g}$ sampled from long replay buffer}
            \State \scalebox{0.9}{$\overline{Q}\leftarrow\max\left(F_\Phi\left(\overline{g}|s',
                        \mu_{\theta}^{\overline{g}}\left(s'\right), \overline{g}\right), 
            Q_{\overline{\omega}}\left(s', \mu^{\overline{g}}_{\theta}\left(s'\right)\right)\right)$}
            \State Update $Q_\omega$ with \scalebox{.85}{$\nabla_\omega \left(r^{\overline{g}}(s, a) + \gamma\overline{Q} - Q_\omega\left(s, a;
            \overline{g}\right)\right)^2$}
            \State Update $\mu^{\overline{g}}_{\theta}$ with \scalebox{0.9}{$\nabla_a Q_\omega\left(s, a;
            \overline{g}\right)\Bigr|_{a=\mu^{\overline{g}}_{\theta}\left(s\right)}\nabla_\theta \mu^{\overline{g}}_{\theta}\left(s\right)$}
        \EndFor
    \EndFor
    \EndFunction
\end{algorithmic}
\caption{Universal Value Density Estimation (UVD)}\label{algorithm:uvd}
\end{algorithm}
\begin{algorithm}[t]
\begin{algorithmic}[1]
    \small
    \Function{VDI}{}
    \For{$i \gets 0..\text{\#Iterations}$}
        \State Fill replay buffer with experience
        \For{$s, a$ sampled from short replay buffer}
            \State Sample time offsets $t\sim \text{Geom}\left(1 - \gamma\right)$
            \State Sample target states $\overline{s}$ $t$ steps ahead of $s$
            \State Update $F_\Phi$ with $-\nabla_\Phi \log F_\Phi\left(\overline{s}|s, a\right)$
            \State Update $d_\Phi$ with $-\nabla_\Phi \log d_\Phi\left(s\right)$
        \EndFor

        \For{$s, a, s'$ sampled from long replay buffer}
                \State Sample $\overline{s}$ uniformly from expert data
                \State $\overline{Q}\leftarrow\max\left(F_\omega \left(\overline{s}|s, a\right), \gamma
                \tilde{Q}_{\overline{\omega}}\left(s', \mu_\theta\left(s'\right)\right)\right)$
                \State Update $\tilde{Q}_\omega$ with $\nabla_\omega \left(\overline{Q} - \tilde{Q}_\omega\left(s, a; \overline{s}\right)\right)^2$
        \EndFor

        \For{$s, a$ from long replay buffer}
            \State Sample $\overline{s}$ from expert data with $p=\frac{1}{d_\Phi(\overline{s})}$
            \State Update $\mu_\theta$ with \scalebox{0.9}{$\nabla_a \tilde{Q}_\omega\left(s, a;
            \overline{s}\right)\Bigr|_{a=\mu_\theta\left(s\right)}\nabla_\theta \mu_\theta\left(s\right)$}
        \EndFor
    \EndFor
    \EndFunction
\end{algorithmic}
\caption{Value Density Imitation (VDI)}\label{algorithm:vdi}
\end{algorithm}
\subsection{Combining Estimators}
Learning a model of $F_\gamma^{\mu_\theta}$ already provides a valid estimator of the Q-function; however, relying on
density estimation alone is insufficient in practice. As the discount factor
approaches $1$, the effective time-horizon becomes large. As a result, estimating Q purely based on density estimation
would require state-action-goal triplets that are hundreds of time-steps apart, leading to updates which suffer from
extremely high variance. Temporal difference learning uses bootstrapping to reduce the variance of the regression
gradient and to handle such large time-horizons. In practice, temporal difference learning is likely to underestimate the
true value as it may never observe a reward\footnote{if the environment is continuous and stochastic and the reward is
as defined above, we can assume that the agent never observes a reward}. Density estimation does not suffer from a
sparse learning signal, but in practice we have to limit the time-horizon to limit variance. The result is that the density estimator will underestimate the value as well. 

Here, we propose to combine both estimators to get the best of both approaches.
There are a variety of ways in which the two estimators can be integrated. \citet{fujimoto2018addressing} combine two
estimators to combat overestimation by choosing the smaller value in the computation of the temporal difference
target. In this work, we follow a similar approach and combat underestimation by choosing the larger of the two
estimators to compute the
temporal difference target. The temporal difference loss is then given by
\begin{align}\label{eq:td_loss}
    &L(\omega) := \left(r^g(s, a) + \gamma \overline{Q} - \tilde{Q}_{\omega}(s, a; g)\right)^2,\\
    &\overline{Q} := \max\left(\tilde{Q}_{\overline{\omega}}(s, a; g), F_\Phi(g|s, a)\right),
\end{align}
where $F_\Phi(g|s, a)$ is a learned estimate of the density defined in Eq.~\ref{eq:density_def}.

\section{Goal-conditioned Reinforcement Learning}\label{sec:uvd_gcrl}
Using the general principle introduced in the previous section, we now introduce Universal Value Density Estimation
(UVD) to address the problem of
goal-conditioned reinforcement learning. Using Eq.~\ref{eq:td_loss}, we can efficiently estimate the Q-function
corresponding to a goal-conditioned reward function;
however, in goal-conditioned reinforcement learning, the policy needs to be conditioned on the
goal as well. Using a goal-conditioned policy in Eq.~\ref{eq:density_def}, we can write down a corresponding equivalence
to a predictive long-term generative model:
\begin{equation}
    F^{\mu^{\overline{g}}}_\gamma (g|s, a) := Q^{\mu^{\overline{g}}}_{r^g}(s, a) 
\end{equation}
Here, $F^{\mu^{\overline{g}}}_\gamma (g|s, a)$ corresponds to the distribution of goals $g$ that the agent will reach
when
it starts in state $s$, takes action $a$ and then tries to reach the goal $\overline{g}$. As before, we can easily
train a model to represent this distribution via density estimation. In this case, the model has to be conditioned on
the intended goal $\overline{g}$ as well and takes the form $F_\Phi(g|s, a, \overline{g})$. Using this density estimate,
we can alter the temporal difference target in Eq.~\ref{eq:td_loss} to efficiently train a UVFA:
\begin{align}\label{eq:td_loss2}
    \overline{Q} := \max\left(Q_{\overline{\omega}}(s, a; g), F_\Phi(g|s, a, g)\right)
\end{align}

We can now write down an iterative algorithm for goal-conditioned reinforcement learning that is able to handle sparse
reward signals while also addressing the problem of hindsight bias. First, the agent collects experience
with a randomly sampled intended goal. By interacting with the environment, the agent observes future achieved goals and
models their distribution conditioned on the intended goal and the current state-action pair. Next, the agent uses the
learned density estimator to update the Q-function as in Eq.~\ref{eq:td_loss2}. Finally, the agent updates the
goal-conditioned policy using the learned UVFA to estimate the policy gradient. The approach is described in more detail in
Algorithm~\ref{algorithm:uvd}, while implementation details are discussed in Appendix~\ref{appendix:practical_considerations}.

\section{Imitation Learning}\label{sec:vdi}
We now turn our attention to the problem of sample-efficient imitation learning. We wish to train the agent to imitate
an expert's policy using only a few demonstration samples from the expert. Different formulations exist to solve this
problem (see Section \ref{sec:imitation}), but the formulation which has arguably been the most promising in the recent past
has been to train the agent to explicitly match the expert's state-action distribution or occupancy 
measure~\citep{ziebart2007maximum, ho2016generative}. Our next step is
therefore to extend our findings from the previous section to
state-action or state distribution matching. 

Similar to \citet{schroecker2017state, schroecker2019generative}, we
propose a maximum-likelihood approach to the distribution-matching problem. Central to this approach is the estimation
of the state-distribution gradient $\nabla_\theta \log d^{\mu_\theta}(\overline{s})$. By itself, following the
state-distribution gradient leads the agent to reproduce the expert's state-distribution which
is ambiguous and may admit multiple solutions. In practice, however, state-distribution matching can often be an effective
approach to imitation learning and is optimal if the expert's behavior can be uniquely described using a reward-function that
depends only on the current state. We will evaluate state-distribution matching as imitation from observation in Section~\ref{sec:exp_imitation}. To use the gradient
estimate for state-action distribution matching, it is possible to combine it with the behavioral cloning gradient as $
    \nabla_\theta \log \rho^{\pi_\theta}(\overline{s}, \overline{a}) = \nabla_\theta \log \pi_\theta(\overline{a}|\overline{s})
+ \nabla_\theta \log d^{\pi_\theta}(\overline{s}),$
where we assume a stochastic policy $\pi_\theta$ in place of a deterministic one. We found, however, that the behavioral cloning gradient can dominate a noisy estimate of the state-distribution gradient
and lead to overfitting. Instead, we propose to
augment the state to include the previous action that lead to the state. This approach attempts to match the joint distribution of action and next state which implies
the state-action-distribution based on environment dynamics.

If applied to a single state, following the state-distribution gradient teaches the agent to go to that state. It is
thus no surprise that it can be shown to be equivalent to the
policy gradient for the right goal-conditioned reward function \citep[also see][]{schroecker2019generative}.
Specifically, the state-distribution gradient is equivalent to the weighted policy gradient in the average-reward
setting (using $r^{\overline{s}}(s,a)=\delta_{s,\overline{s}}$). We have:
\begin{equation}
\begin{aligned}
    \nabla_\theta \log d^{\mu_\theta}(\overline{s}) &=
    \frac{\nabla_\theta d^{\mu_\theta}(\overline{s})}{d^{\mu_\theta}(\overline{s})}\\ 
    &= \nabla_\theta \frac{\int d^{\mu_\theta}(s) \delta_{s, \overline{s}}ds}{d^{\mu_{\theta}}(\overline{s})}  
    = \frac{\nabla_\theta J^{r^{\overline{s}}}(\theta)}{{d^{\mu_{\theta}}(\overline{s})}}
\end{aligned}
\end{equation}
Intuitively, the policy gradient leads the agent toward a demonstration state $\overline{s}$, while the weight ensures
that all demonstration states are visited with equal probability.  This gives rise to Value Density Imitation Learning
(see Alg. \ref{algorithm:vdi}): 
    \begin{enumerate}
        \item Using self-supervised roll-outs, learn the goal-conditioned Q function as in section \ref{sec:uvd}.
        \item Using same roll-outs, train an unconditional density estimator to model the agent's state-distribution:
            $d_\omega(s)$.
        \item Sample demonstration states with probability proportional to $\frac{1}{{d_\omega(\overline{s})}}$.
        \item Use the learned Q function, conditioned on the sampled demonstration states, to estimate the policy
            gradient.
    \end{enumerate}
\section{Experiments}
\begin{figure*}[t]
    \centering
    \begin{subfigure}{0.246\textwidth}
        \includegraphics[width=\textwidth]{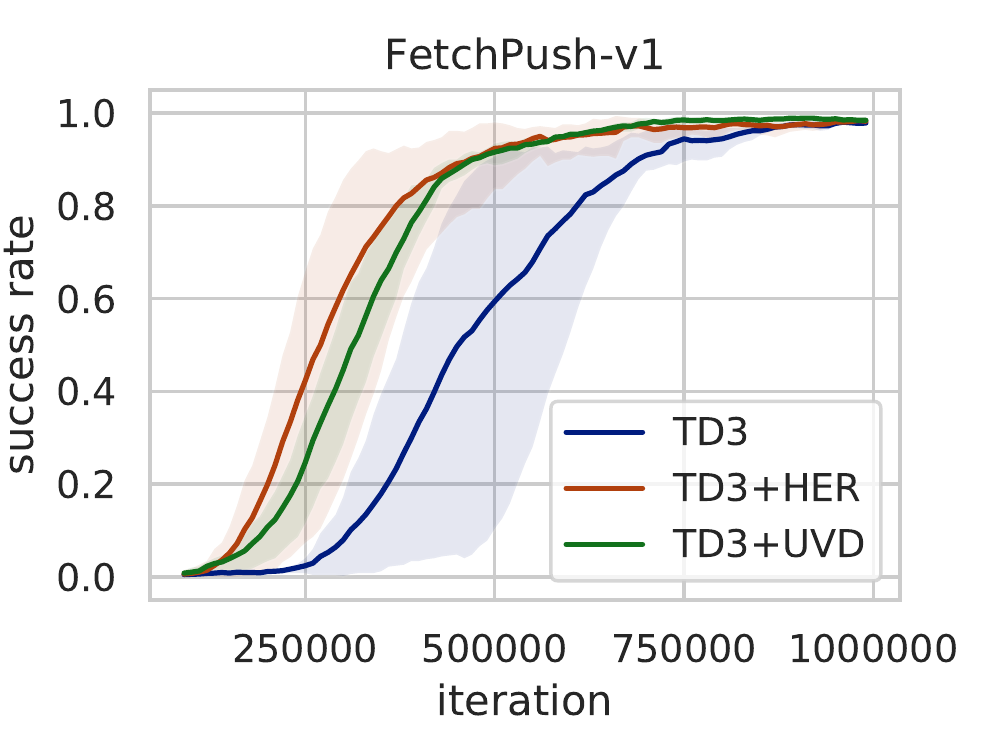}
        \caption{}\label{fig:fetch_push}
    \end{subfigure}
    \begin{subfigure}{0.246\textwidth}
        \includegraphics[width=\textwidth]{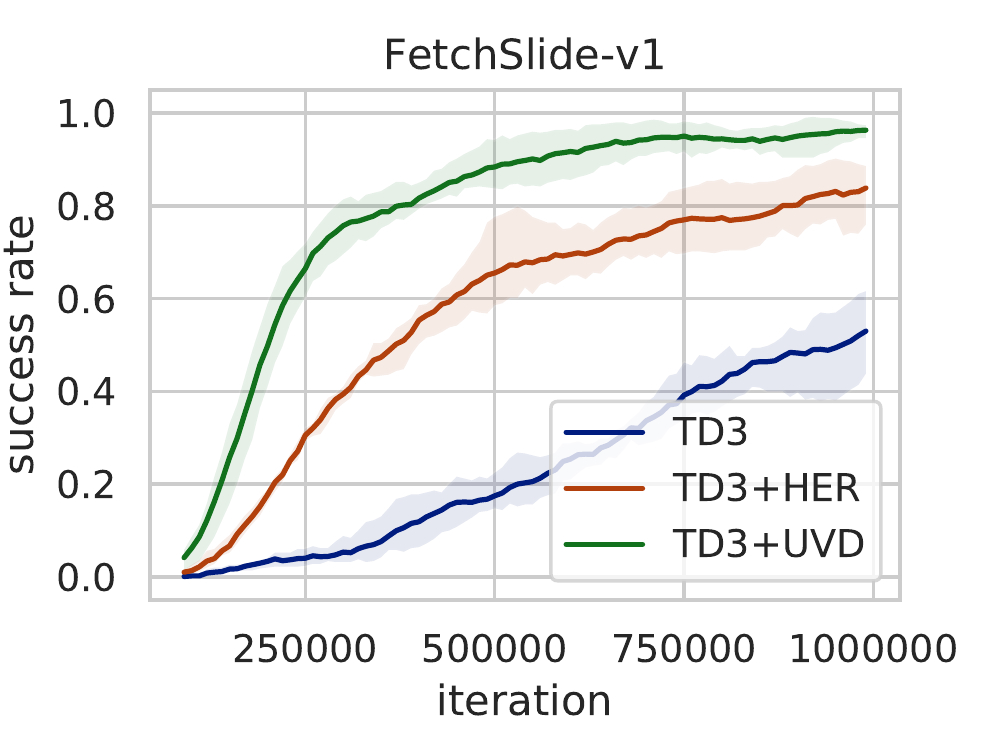}
        \caption{}\label{fig:fetch_slide}
    \end{subfigure}
    \centering
    \begin{subfigure}{0.246\textwidth}
        \includegraphics[width=\textwidth]{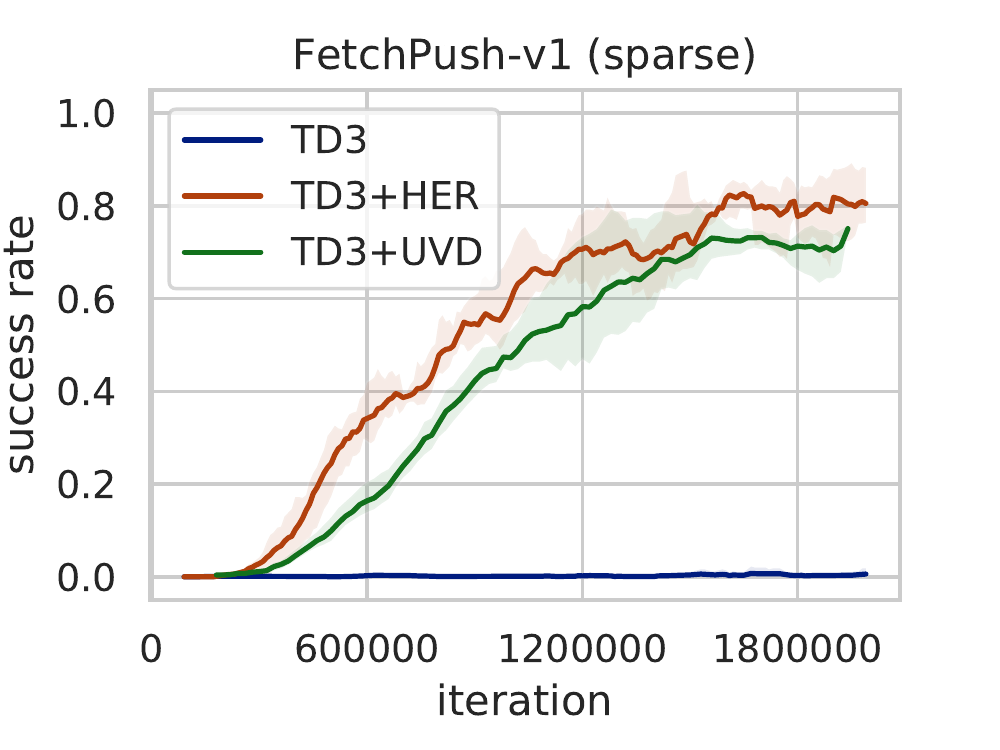}
        \caption{}\label{fig:difficult_push}
    \end{subfigure}
    \begin{subfigure}{0.246\textwidth}
        \includegraphics[width=\textwidth]{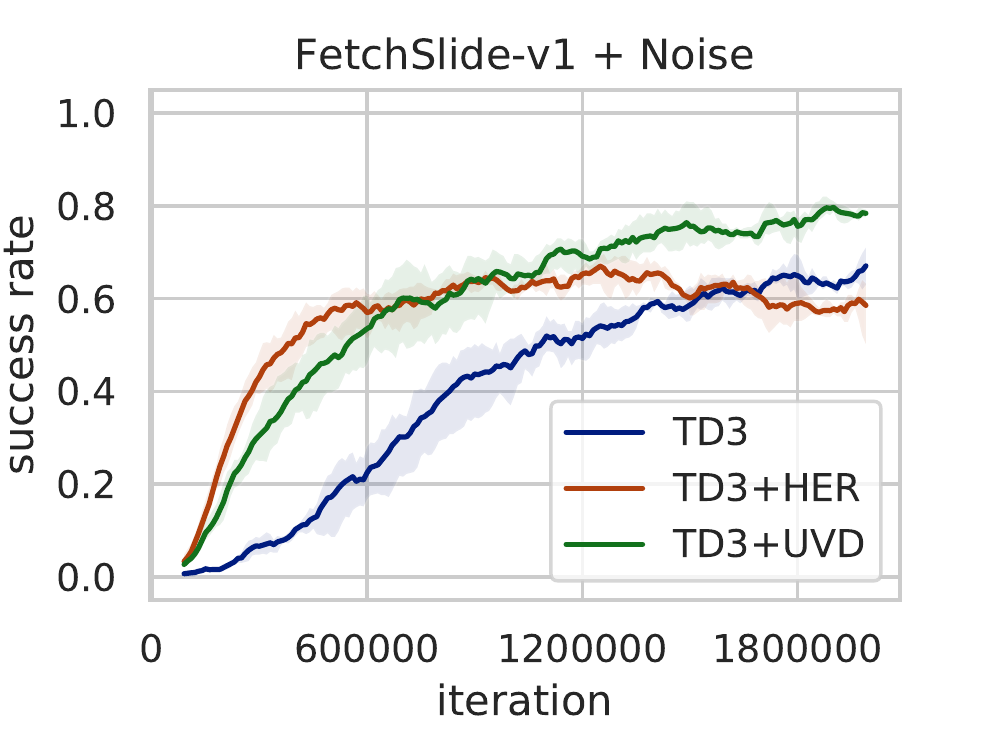}
        \caption{}\label{fig:noisy_slide}
    \end{subfigure}
    \caption{Comparison of TD3, UVD and HER on variations of the Fetch manipulation domains. \emph{FetchPush} and
        \emph{FetchSlide} correspond to the original domains. \emph{FetchPush (Sparse)}
    shows the advantage of hindsight updates when the required precision is significantly higher while
\emph{FetchSlide + Noise} shows the effect of hindsight bias in stochastic domains.}
\end{figure*}
\begin{figure*}[t]
    \centering
    \begin{subfigure}{0.45\textwidth}
        \includegraphics[width=\textwidth]{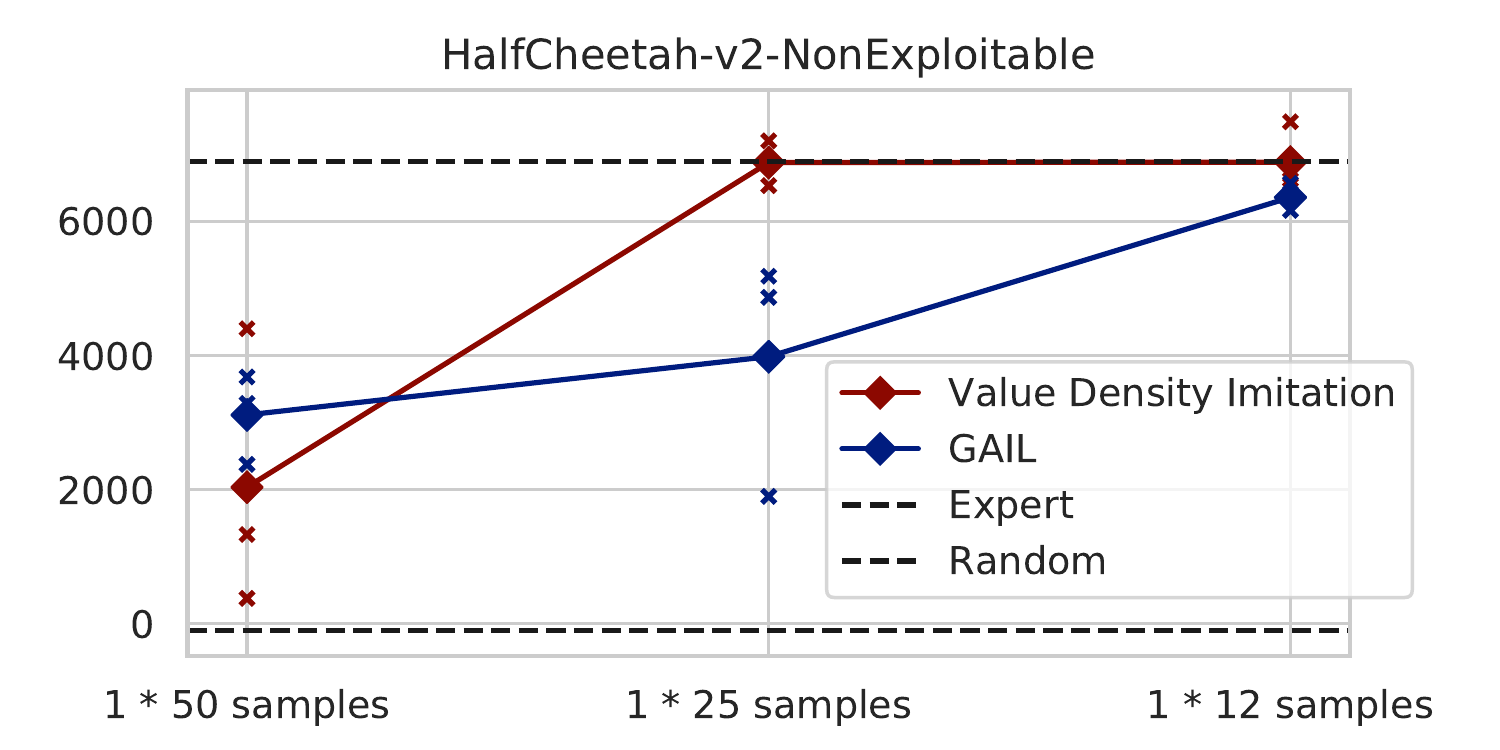}
        \caption{}\label{fig:cheetah}
    \end{subfigure}
    \begin{subfigure}{0.45\textwidth}
        \includegraphics[width=\textwidth]{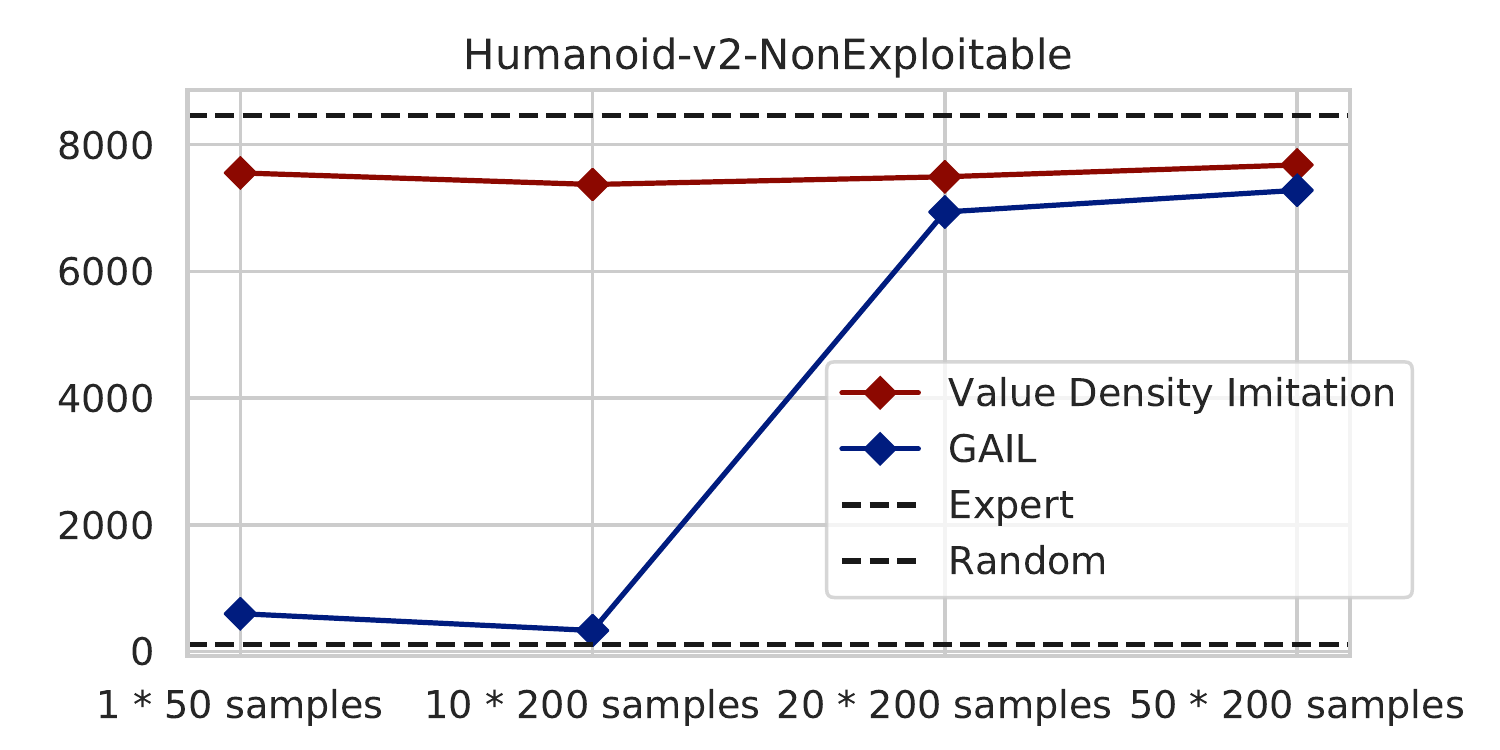}
        \caption{}\label{fig:humanoid}
    \end{subfigure}
    \caption{Comparison of GAIL and VDI on variations of standard benchmark tasks. x-axis shows the number of provided demonstration trajectories as well as the number of state-action pairs collected from each trajectory. VDI is able to
    outperform GAIL when the number of provided demonstrations is low.}\label{fig:imitation_eval}
\end{figure*}
\subsection{Goal-Conditioned Reinforcement Learning}\label{sec:exp_gcrl}
We first evaluate UVD on a suite of simulated manipulation tasks involving a Fetch robot arm. This suite has been developed by \citet{andrychowicz2017hindsight} to show
the strengths of hindsight experience replay. As the simulation is fully deterministic, the effects of hindsight bias
don't play a role in these environments. Our goal here is therefore to show UVD to
solve the unmodified tasks as efficiently as HER while outperforming it on a stochastic variation of the domain. We
compare all methods using the same hyper-parameters, which can be found
in Appendix \ref{sec:appendix_fetch_hypers}.

The first domain in the suite, \emph{FetchPush}, requires the robot arm to learn to push an object to any given target
location. The reward signal is sparse as the agent is
given a non-zero reward only if the object reaches the desired location. In Figure \ref{fig:fetch_push}, we can see that  TD3 with
HER and TD3 with UVD perform similarly. Contrary to the original findings by
\citet{andrychowicz2017hindsight}, we also find unmodified TD3 to be able to solve the task accurately using roughly
twice the amount of training samples. This indicates that the area around the goal in which the object is
considered to be at the desired location is relatively large. If we apply a stricter criterion for the goal being
reached by reducing the size of the goal area by a factor of 100, we can see that hindsight samples are
necessary to learn from sparse rewards. In this variant, TD3 fails to learn a useful policy while both UVD and HER performing similarly (see Figure
    \ref{fig:difficult_push}).

The second domain in the suite, \emph{FetchSlide}, requires the robot to slide the object toward a desired location that is
out of reach of the robot arm. In Figure \ref{fig:fetch_slide}, we can see UVD and HER learning to solve the task
quickly while TD3 without hindsight samples requires significantly more training samples. Unlike in the case of
\emph{FetchPush}, TD3+UVD learns slightly faster in this domain than TD3+HER but both are able to solve the task
eventually.
Both, FetchSlide and FetchPush, are deterministic domains that play to the strengths of hindsight experience replay. In
practice, however, manipulation with a real robot arm is always noisy. In some cases, HER can overcome this noise
despite suffering from hindsight bias; however, this is not always the case. Here, we introduce a variation of the
FetchSlide domain which adds Gaussian noise to the actions of the
agent. We scale this noise based on the squared norm of the chosen actions $\frac{1}{2e}||\max(0,
a-0.5\cdot\mathbf{1})||_2^2$. This scaling allows the agent to adapt to the noise; however, doing so requires the agent to accurately assess the risks of its actions. 
In Figure \ref{fig:noisy_slide}, we can
see that while HER initially learns quickly as in the deterministic domain, it converges to a policy that is noticeably
worse than the policy found by TD3+UVD. This shows the effect of hindsight bias: in the presence of noise, the agent
assumes the noise to be benign. UVD, on the other hand, estimates the risk accurately and achieves a higher success rate
than HER.
\subsection{Imitation Learning}\label{sec:exp_imitation}
Next, we evaluate Value Density Imitation on common benchmark tasks and show that it outperforms the current state of
the art. We point out issues with using these domains in an imitation learning context verbatim and introduce more difficult variations of the tasks that address those issues. The suite of simulated locomotion tasks found in OpenAI Gym~\citep{brockman2016openai} has become a standard
benchmark task for reinforcement learning due to their complex dynamics and proprioceptive,
relatively low-dimensional state-action spaces. Recently, the suite of benchmark tasks has been used to evaluate
imitation learning algorithms as well with~\citet{ho2016generative} showing GAIL to solve the tasks using only a handful
of demonstrated trajectories; however, we find that the unmodified locomotion tasks, despite their popularity, are
easily exploited in an imitation learning context. While it is impressive that GAIL is able to solve \emph{Humanoid-v2} with less than a dozen of disjointed
demonstrated states, it is also clear that the agent is not learning to imitate the motion itself. 

The dominating source of reward in these tasks comes from the velocity in a particular
direction. This is problematic as the
velocity is fully observable as part of the state and, in the case of humanoid, may be encoded in more than one of the features found in the
state-vector. Even the simple average of the state-features may be equivalent to a noisy version of the original reward signal.
Moreover, as the reward is a linear combination of state-features, we know that accurate distribution matching is not
necessary and matching feature expectations is sufficient~\citep{abbeel2004apprenticeship,ho2016generative}. To
alleviate this, we remove task-space velocities in $x,y$ directions from the state-space. A second source of bias can be found in the termination condition of the locomotion
domains. \citet{kostrikov2018discriminator} point out that GAIL is biased toward longer trajectories and thus tries to avoid termination, which in the case of locomotion means to avoid falling. While \citeauthor{kostrikov2018discriminator} adjust the algorithm itself to avoid such
bias, we instead propose to remove the termination condition and use an evaluation which cannot be exploited by a
biased method. Besides removing bias, this makes the learning problem significantly more difficult as a significant
portion of the agent's experience is far from the demonstrated path.

We focus on two locomotion tasks in particular: \emph{HalfCheetah-v2} and
\emph{Humanoid-v2}.
We choose \emph{HalfCheetah-v2} as it is comparatively easy to train an agent to move in the right direction while it is
comparatively difficult to move at high speeds. With the original threshold for solving the task being set at a score of
4500, recent advances in reinforcement learning train policies that achieve 3-4 times as high a
score~\citep{fujimoto2018addressing,haarnoja2018soft} (although we find that removing velocity from the state reduces
    the top-speed achieved by the TD3-trained expert).
    Our second domain of choice is Humanoid-v2, which is generally considered to be the most complex locomotion task. Unlike
in the \emph{HalfCheetah-v2} domain, learning to move without falling can be a significant challenge for a learning agent.
We furthermore find it sufficient to match state-distributions to solve \emph{HalfCheetah-v2} and thus teach the agent
from observation only when using VDI. In the case \emph{Humanoid-v2}, we find that demonstrated actions significantly aid
exploration and thus include them.
We compare the demonstration-efficiency of Value Density Imitation with GAIL, the 
state-of-the-art in terms of demonstration-efficiency on these domains. Similar to \citet{ho2016generative}, we
subsample each trajectory to make the problem more challenging. 
The results can be seen in Figures
\ref{fig:cheetah} and \ref{fig:humanoid}. While both methods are able to achieve near-expert
performance on \emph{HalfCheetah-v2-NonExploitable} using a single demonstrated trajectory sub-sampled at the same rate
as used by \citeauthor{ho2016generative}, we find VDI to be able to imitate the expert if the trajectory is
sub-sampled even further. On \emph{Humanoid-v2-NonExploitable}, we find the difference to be more drastic: while GAIL is
able to learn locomotion behavior from a similar number of trajectories as used in the original paper (but using more
state-action pairs), the performance drops off quickly if we reduce the number of trajectories further. VDI is able to
imitate the expert using only a single demonstrated trajectory. 

Both methods are able to achieve great demonstration-efficiency by matching the expert's state-action distribution;
however, GAIL does so by learning a distance function using demonstrations as training samples. As the number of
demonstrations shrinks, learning a good discriminator to serve as a reward becomes more difficult. VDI side-steps this
issue by learning a goal-conditioned Q-function based on self-supervised roll-outs alone. The demonstrations are used to
condition the Q-function and don't serve directly as training data for any network.
\section{Related Work}

Addressing the problem of goal-conditioned reinforcement learning, 
\citet{sutton2011horde} introduce HORDE, the first paper to utilize recent advances in off-policy RL to learn
multiple value functions simultaneously and the foundation for UVFA~\citep{schaul2015universal} and
HER~\citep{andrychowicz2017hindsight} which our work heavily relies on. 
Recent work by \citet{nair2018visual} and \citet{sahni2019visual} propose approaches that adapt HER to the setting of high-dimensional, visual features. While our work is based on RealNVP which was originally invented for modeling images, learning universal value densities in such a high-dimensional setting comes with its own set of challenges, not the least of which are computational, and is an interesting avenue for future research.
To the best of our knowledge, the first paper to
identify bias in HER is \citep{lanka2018archer}; however, while the authors propose a heuristic method to handle this issue, no
principled method to use hindsight samples in a completely unbiased way has been proposed to date. Most closely related
to the method of Universal Value Density Estimation itself are Temporal Difference Models~\citep{pong2018temporal}. TDMs use a
squared error function as a terminal value and use bootstrapping to propagate this value over a finite horizon.
$F_\gamma$ can similarly be seen as a model of terminal value, where the use of it as a lower bound allows bootstrapping
with an infinite time-horizon. While there are further similarities to our method when using a Gaussian density estimator, the proposed method does not address hindsight bias, TDMs do not address hindsight bias.

Value Density Imitation is closely related to other work in state-action-distribution matching such as
adversarial methods. \citet{ho2016generative} show early work in inverse reinforcement learning to be reducible to
state-action distribution matching~\citep[e.g.][]{abbeel2004apprenticeship,ziebart2007maximum} and furthermore introduce
GAIL. Using an adversarial objective, GAIL trains a discriminator to identify expert transitions while using
reinforcement learning to train the agent to fool the discriminator. Several approaches have been
introduced that build on this idea. Adversarial Inverse Reinforcement Learning uses an adversarial objective to learn a
fixed reward function~\citep{fu2018learning} while Discriminator Actor-Critic~\citep{kostrikov2018discriminator}
reduces the number of environment interactions required. GPRIL~\citep{schroecker2019generative} attempts to match the
state-action-distribution using a similar long-term generative model. In contrast to VDI, GPRIL cannot easily be
combined with temporal-difference learning techniques as the model is used to generate samples and struggles with larger
time-horizons as can be found in the locomotion benchmark tasks. 

\bibliography{library}
\bibliographystyle{apa}
\appendix
\section{Practical considerations}\label{appendix:practical_considerations}
There are a number of implementations decisions that were made to improve the sample-efficiency and stability of Value
Density Estimation and its application to imitation learning. Here, we review these decisions in more detail.
\subsection{Universal Value Density Estimation}
\paragraph{Using an exploration policy:}In most cases self-supervised roll-outs will require the agent to explore. In
our method, we combine a temporal difference update rule as is usually found in deterministic policy gradients with
density estimation. While the temporal-difference update rule can handle off-policy data from an exploration policy,
density-estimation is on-policy. In practice, however, \citet{fujimoto2018addressing} add Gaussian noise to the target-Q function and report
better results by learning a smoothed Q-function that is akin to the on-policy Q-function with Gaussian exploration
noise. It thus stands to reason that we can omit off-policy correction in the density-estimator.
\paragraph{Truncating the time horizon:}The training data for learning a long-term model can be fairly noisy. While we
can expect density estimation to be efficient over a horizon of just a few time-steps, the variance increases
dramatically as $\gamma$ increases. This is the primary motivation for utilizing temporal-difference learning in
conjunction with universal value density estimation. To better facilitate stable training, we truncate the time-horizon
of the density-estimator to a fixed number of time-steps $T$. The temporal-difference learning component is thus solely
responsible for propagating the value beyond this fixed horizon. The effect this has on the optimal policy is small: the
Q-value will be underestimated by ignoring visitations with time-to-recurrence greater than $T$. A greater time-horizon
boosts the effect of hindsight samples and leads to a learning signal that is less sparse while a smaller time-horizon
reduces the variance of the estimator.
\paragraph{Using a replay buffer:}Using a replay buffer is essential for sample-efficient training with model-free
reinforcement- and imitation-learning methods and improves the stability of deterministic policy gradients. We find that
this is true for training long-term generative models as well. While importance sampling based off-policy correction for density estimation is possible, we find that it introduces instabilities and is thus undesirable. Instead, we propose to use a separate, shorter replay-buffer for density estimation to mitigate the undesirable effects of off-policy learning while retaining some of the benefits.
\paragraph{Delayed density updates:}Temporal-difference learning with non-linear function approximation is notoriously
unstable. To help stabilize it, a common~\citep{mnih2015human,fujimoto2018addressing} trick is delay the update of the target network and allow the
Q-function to perform multiple steps of regression toward a fixed target. Since we use the long-term predictive model
$F_\omega$ to calculate the temporal-difference regression target, we apply the same trick here. We maintain a target
network $F_{\overline{\omega}}$ which we set to be equal to the online density estimator $F_\omega$ after a fixed number of
iterations. In Value Density Imitation, we use the same procedure to maintain a frozen target network of the
unconditional state density estimator $d_\omega$.
\paragraph{Normalizing states:}As our method depends on density estimation, the resulting values are heavily affected by
the scale of the features. We therefore normalize our data based on the range observed in random roll-outs as well as, in
the case of imitation learning, the range seen in the given demonstrations.
\subsection{Value Density Imitation}
\paragraph{Averaging logits:} While the dimensionality of the goal in goal-conditioned reinforcement learning is
typically small, Value Density Imitation requires us to use the entire state as a goal. This, however, can be difficult
if the number of features is large. If the state-features are independent, the density suffers from the curse of
dimensionality as it is multiplicative and the Q-values will be either extremely large or extremely small. Even if the
true density function does not exhibit this property, Normalizing Flows predict the density as a product of $N$
predicted logits and prediction errors are therefore multiplicative. To combat this, we take the average of the
predicted logits rather than the sum, effectively taking the $N-th$ square root of the Q-function. We find that this
approximation works well in practice and justify it further in appendix \ref{sec:appendix_dimensionality}.
\paragraph{Bounding weights on demonstrated states:} In Value Density Imitation, we sub-sample demonstrations states
proportional to $\frac{1}{d_\omega(\overline{s})}$ to ensure demonstration states to be visited with equal probability. In this
formulation, demonstration states that are especially difficult to reach may be over-sampled by a large factor and
destabilize the learning process. To counteract this, we put a bound on the weight of each demonstration state:
for each batch, the weights are normalized and an upper bound is applied.
\paragraph{Spatial and temporal smoothing:} We apply two kinds of smoothing to the learned UVFA to improve the stability
of the learning algorithm. Spatial smoothing ensures that similar state-action pairs have similar value and is achieved
by adding Gaussian noise to training samples of the target state when training the density estimator. Temporal smoothing
ensures that the learned value does not spike too suddenly. Instead of using $F_\gamma(\overline{s})$ as a temporal
difference regression target, we use a mixture of the density estimation and the temporal-difference lower bound. Using
a temporal smoothing factor $\lambda$, the full temporal difference loss is then given by:
\begin{align*}
    L(\omega) :=& \left(r^g(s, a) + \gamma\overline{Q} - \tilde{Q}_{\omega}(s, a; g)\right)^2,\\
    \overline{Q} :=& \lambda\tilde{Q}_{\overline{\omega}}(s, a; g) + \\&(1-\lambda)\max\left(\tilde{Q}_{\overline{\omega}}(s, a; g), F_\Phi(g|s, a)\right)
\end{align*}

\section{Escaping the curse of dimensionality}\label{sec:appendix_dimensionality}
In section \ref{sec:uvd}, we introduced a method which uses the probability density predicted by a normalizing flow as a Q function. We showed that this density Q function is a valid Q function with a corresponding reward function that is sensible for many practical task. In this appendix, we consider the numerical properties of the universal value density estimator and propose a slight variation that is easier to handle numerically. To see the numerical challenge in using density estimators as value functions, we take a look another look at a single bijector of a RealNVP; here, the bijector $f_\omega(z)=\left(s_\omega(z), t_\omega(z)\right)$ is predicting an affine transformation of $z\sim p_z(\cdot) = \mathcal{N}(0, I)$ to $x\sim p_x(\cdot)$, i.e. $x_i = s_\omega(z)_i z_i + t_\omega(z)_i$. The log-density of x is then given as $p_x(x) = e^{\sum_{i=0}^N s^{-1}_\omega(x) - \left(\frac{x_i - t^{-1}_\omega(x)_i }{s^{-1}_\omega(x)_i}\right)^2 + \log \frac{1}{\sqrt{2\pi}}}$.  
It is readily apparent that this value can easily explode, especially when used as a target Q value in the mean-squared
loss of a temporal difference update. To combat this, we propose to scale the logits with the dimensionality $N$, i.e.
we use $p_x(x) = e^{\frac{1}{N}\sum_{i=0}^N s^{-1}_\omega(x) - \left(\frac{x_i - t^{-1}_\omega(x)_i
}{s^{-1}_\omega(x)_i}\right)^2 + \log \frac{1}{\sqrt{2\pi}}}$. As this corresponds to only a constant factor on $\log
F_\gamma$, the gradient-based density estimation is not affected. 

While the change is simple, average logits instead of taking the sum, we need to justify the approximation anywhere the
value density and the Q-function are used: first, we
show that $J(\theta)^\frac{1}{N}$ can be used in place of $J(\theta)$ in both, goal-conditioned reinforcement learning
and in imitation learning without changing the optimal policy; second, we justify using $Q^{\mu_\theta}_{r^g}(s,
a)^\frac{1}{N}$ in place of $Q^{\mu_\theta}_{r^g}(s, a)$ when computing the policy gradient; and, finally, we
show that we can justify the use of the N-th root in the temporal difference learning update.

\paragraph{Using the scaled objective $J(\theta)^\frac{1}{N}$:} Here, we have to consider two cases. In the case of
goal-conditioned RL, we have to show that $\max_\theta J(\theta) = \max_\theta J(\theta)^\frac{1}{N}$. To this end, it
is sufficient to note that the reward function is strictly non-negative and thus $\left(\cdot\right)^\frac{1}{N}$ is a
monotonous function. In the case of imitation learning, we can immediately see that the change corresponds to a constant factor on
the state-distribution gradient:
\begin{equation}
    \frac{1}{N} \nabla_\theta \log d^{\mu_\theta}(\overline{s}) = \frac{\nabla_\theta
    J_{r^{\overline{s}}}(\theta)^{\frac{1}{N}}}{{d^{\mu_{\theta}}(\overline{s})^{\frac{1}{N}}}}
\end{equation}

\paragraph{Estimating the policy gradient $\nabla_\theta J(\theta)^\frac{1}{N}$:} The deterministic policy gradient
theorem~\citep{silver2014deterministic} shows that maximizing the Q-value in states sampled from the agent's discounted on-policy
state-distribution is equivalent to maximizing the reinforcement-learning objective. This is immediately apparent if the
representation of the policy is sufficiently expressive and agent is able to take the action with maximum value in every
state. Due to monotonicity of $\left(\cdot\right)^\frac{1}{N}$, this is true when using $Q_\Phi(s, a; g)^\frac{1}{N}$ as well.
If the policy is not able to maximize the Q-function everywhere, the deterministic policy gradient theorem shows that
sampling from the discounted state-distribution leads to the agent making the right trade-offs. Using $Q(s, a;
g)^\frac{1}{N}$, however, leads to a different trade-off. In practice, this is typically ignored: deterministic policy
gradients used off-policy with a replay-buffer are not guaranteed to make the right trade-off and even if they are used
on-policy, the discount-factor is typically ignored and the resulting estimate of the policy gradient is
biased~\citep{nota2019policy}.
\paragraph{$Q^{\mu}_g(s, a)^\frac{1}{N}$ as TD target:} 
Finally, we need to show that we can use the N-th root in the computation of the temporal difference learning target. To this end, we make
use of the fact that the reward in continuous environments can be assumed to be 0 when computing the temporal difference
target. In this case $\left(r^g(s,a) + \gamma Q^\mu_{r^g}(s', \mu(s'))\right)^{\frac{1}{N}}$ becomes $\gamma^\frac{1}{N}
Q^\mu_{r^g}(s', \mu(s'))^{\frac{1}{N}}$ and increasing $\gamma$ is sufficient to compute a valid regression target
for $Q^\mu_{r^g}(s, a)^{\frac{1}{N}}$
\section{Hyperparameters}\label{sec:appendix_fetch_hypers}\label{sec:appendix_locomotion_hypers}
\begin{table*}[t]
\centering \scriptsize
    \begin{tabular}{|l|l|}
        \hline
        \multicolumn{2}{|l|}{\bf General parameters} \\\hline
        Environment steps per iteration & 1 \\\hline 
        $\gamma$ & 0.98 \\\hline
        Batch size & 512\\\hline
        Replay memory size & 1500000 \\\hline
        Short replay memory size & 50000 \\\hline
        Sequence Truncation (Density estimation training) & 4 \\\hline
        Optimizer & Adam\\\hline
        \multicolumn{2}{|l|}{\bf Policy} \\\hline
        Hidden layers & 400, 400\\\hline
        Hidden activation & leaky relu\\\hline
        Output activation & tanh\\\hline
        Exploration noise $\sigma$ & 0.1 \\\hline
        Target action noise $\sigma$ & 0.0 \\\hline
        Learning rate & $2\cdot 10^{-4}$ (TD3, TD3+HER), $8\cdot 10^{-4}$ (TD3+UVD) \\\hline
        \multicolumn{2}{|l|}{\bf Q-network} \\\hline
        Hidden layers & 400, 400\\\hline
        Hidden activation & leaky relu\\\hline
        Output activation & 50 tanh (TD3, TD3+HER), linear (TD3+UVD)\\\hline
        Learning rate & $2\cdot 10^{-4}$ (TD3, TD3+HER), $8\cdot 10^{-4}$ (TD3+UVD) \\\hline
        \multicolumn{2}{|l|}{\bf RealNVP} \\\hline
        Bijector hidden layers & 300, 300\\\hline
        Hidden activation & leaky relu\\\hline
        Output activation, scale & tanh(log())\\\hline
        Output activation, translate & linear\\\hline
        num bijectors & 5 (slide), 6 (push)\\\hline
        Learning rate & $2 \cdot 10^{-4}$\\\hline
    \end{tabular}
    \caption{Common parameters in Fetch environments}\label{fig:uvd_params}
\end{table*}
In Table \ref{fig:uvd_params}, we list the hyper-parameters used for TD3, TD3+HER and TD3+UVD on the Fetch experiments. When applicable, we
largely use the same hyper-parameters for each algorithm. Two important exceptions are the learning rate which is tuned
individually for each algorithm (TD3+UVD benefits reliably from higher learning rates whereas TD3 and TD3+HER does not always
converge to a good solution at higher learning rates) as well as the output activation of the Q-network. A tanh
activation is used to scale the value to the range of -50 to 50 in the case of TD3 and TD3+HER as we found this to
drastically improve performance. Since the density is not bounded from above, the same activation cannot be used in the
case of TD3+UVD.

In Table \ref{fig:vdi_params}, we list the hyper-parameters used for VDI in the locomotion experiments. Parameters are largely identical between
environments; however, in some cases we trade off higher learning speed for reduced accuracy on \emph{HalfCheetah}. 
In the case of GAIL, we use the implementation found in OpenAI baselines\footnote{https://github.com/openai/baselines/tree/master/baselines/gail}, using 16 parallel processes (collecting 16 trajectories per iteration) and do not modify the parameters.
\begin{table*}[t]
\centering \scriptsize
    \begin{tabular}{|l|l|l|}
        \hline
        Parameter & \emph{HalfCheetah} & \emph{Humanoid}\\\hline
        \multicolumn{3}{|l|}{\bf General parameters} \\\hline
        Environment steps per iteration & 10 & 10\\\hline 
        $\gamma$ & 0.995 & 0.995 \\\hline
        Batch size & 256 & 256\\\hline
        Replay memory size & 1500000 & 1500000\\\hline
        Short replay memory size & 500000 & 500000 \\\hline
        Sequence Truncation (Density estimation training) & 4 & 4 \\\hline
        Optimizer & Adam & Adam\\\hline
        \multicolumn{3}{|l|}{\bf Policy} \\\hline
        Hidden layers & 400, 300 & 400, 300 \\\hline
        Hidden activation & leaky relu & leaky relu\\\hline
        Output activation & tanh & tanh\\\hline
        Exploration noise $\sigma$ & 0.3 (until iteration 25000), 0.1 (after) & 0.1 \\\hline
        Target action noise $\sigma$ & 0.0 & 0.0 \\\hline
        Learning rate & $3\cdot 10^{-4}$ & $3\cdot 10^{-4}$ \\\hline
        \multicolumn{3}{|l|}{\bf Q-network} \\\hline
        Hidden layers & 400, 400 & 400, 400\\\hline
        Hidden activation & leaky relu & leaky relu\\\hline
        Output activation & linear & linear\\\hline
        Learning rate & $3\cdot 10^{-4}$ & $1\cdot 10{-4}$\\\hline
        \multicolumn{3}{|l|}{\bf RealNVP} \\\hline
        Bijector hidden layers & 400, 400 & 400, 400\\\hline
        Hidden activation & leaky relu & leaky relu\\\hline
        Output activation, scale & tanh(log()) & tanh(log())\\\hline
        Output activation, translate & linear & linear\\\hline
        num bijectors & 5 & 5\\\hline
        Learning rate & $1\cdot 10^{-4}$ & $2\cdot 10^{-5}$\\\hline
        L2-regularization & $1\cdot 10^{-5}$ & $1\cdot 10^{-6}$\\\hline
        Spatial smoothing & 0.1 &0.1\\\hline
        Temporal smoothing & 0. & 0.98\\\hline
    \end{tabular}
    \caption{VDI parameters in locomotion environments}\label{fig:vdi_params}

\end{table*}
\end{document}